\documentclass[letterpaper, 10 pt, conference]{ieeeconf}  

\IEEEoverridecommandlockouts                              

\usepackage{graphics} 
\usepackage{epsfig} 
\usepackage{times} 
\usepackage{amsmath} 
\usepackage{amssymb}  
\usepackage{adjustbox}
\usepackage{multirow}

\usepackage{multicol}
\usepackage[bookmarks=true]{hyperref}
\usepackage[usenames,dvipsnames]{xcolor}

\title{\LARGE \bf
Grasping Trajectory Optimization with Point Clouds
}

\author{Yu Xiang, Sai Haneesh Allu, Rohith Peddi, Tyler Summers, Vibhav Gogate
\thanks{Yu Xiang, Sai Haneesh Allu, Rohith Peddi and Vibhav Gogate are with the Department of Computer Science, and Tyler Summers is with the Department of Mechanical Engineering, University of Texas at Dallas, Richardson, TX 75080, USA \tt\small \{yu.xiang, saiHaneesh.allu, rohith.peddi, tyler.summers, vibhav.gogate\}@utdallas.edu}
}

\begin{document}

\maketitle
\thispagestyle{empty}
\pagestyle{empty}

\begin{abstract}

We introduce a new trajectory optimization method for robotic grasping based on a point-cloud representation of robots and task spaces. In our method, robots are represented by 3D points on their link surfaces. The task space of a robot is represented by a point cloud that can be obtained from depth sensors. Using the point-cloud representation, goal reaching in grasping can be formulated as point matching, while collision avoidance can be efficiently achieved by querying the signed distance values of the robot points in the signed distance field of the scene points. Consequently, a constrained nonlinear optimization problem is formulated to solve the joint motion and grasp planning problem. The advantage of our method is that the point-cloud representation is general to be used with any robot in any environment. We demonstrate the effectiveness of our method by performing experiments on a tabletop scene and a shelf scene for grasping with a Fetch mobile manipulator and a Franka Panda arm. \footnote{Code and videos for the project are available at \\ \url{https://irvlutd.github.io/GraspTrajOpt}}

\end{abstract}

\section{Introduction}

In robot manipulation, planning a robot trajectory to grasp an object is a fundamental research problem. The problem is challenging since it requires motion planning to avoid obstacles in the task space and grasp planning to decide how to grasp a target object. Traditionally, the motion planning problem and the grasp planning problem are tackled separately. Motion planning approaches focus on finding a collision-free path to reach a given end-effector goal. For example, sampling-based motion planning methods such as Rapidly exploring Random Trees (RRTs) \cite{lavalle1998rapidly,kuffner2000rrt,karaman2011sampling} and Fast Marching Tree (FMT) \cite{janson2015fast} find robot trajectories by incrementally building configuration space filling trees through directed sampling. Optimization-based motion planning methods~\cite{ratliff2009chomp,schulman2014motion,mukadam2018continuous,stouraitis2020multi} solve optimization problems to find robot trajectories that minimize some loss functions and obey certain constraints, such as joint limits.

Since these motion planning algorithms need to have a given goal, they cannot be applied directly to robot grasping unless a grasping goal is given. On the other hand, grasp planning methods such as GraspIt!~\cite{miller2004graspit}, 6D GraspNet~\cite{mousavian20196} and SE(3)-DiffusionFields~\cite{urain2023se} aim to synthesize grasps of robot grippers given 3D models or 3D point clouds of objects. These methods focus on planning the poses of robot grippers to grasp various objects. However, they do not consider the motion of the robotic arm to reach the planned grasps.

Combining grasp planning and motion planning can address the robot grasping problem. A straightforward approach is first to utilize a grasp planning method to generate grasps of a target object and then employ a motion planning method to plan a robot trajectory to reach one of the grasps. A naive way is to loop over all the planned grasps until the motion planner finds a collision-free path to reach one of the grasps. This naive approach is complete, that is, as long as there is one plausible grasp from the grasp planner, the method can find a path to reach it. However, it is very slow, especially when the number of planned grasps is large. Therefore, a number of approaches are proposed to address the problem of joint motion and grasp planning.

\begin{figure} 
\centering
\begin{center} 
\includegraphics[width=0.48\textwidth]{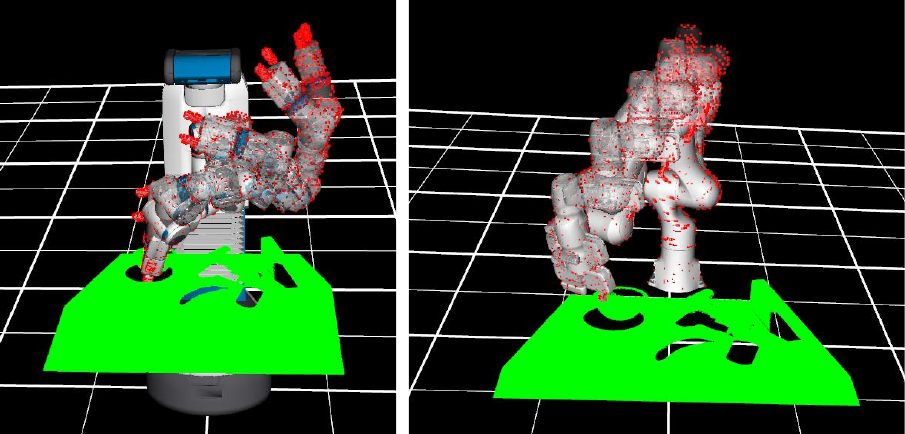}
\caption{We represent robots and the task space with point clouds, and solve a trajectory optimization problem for joint motion and grasp planning.}
\label{fig:intro}
\vspace{-6mm}
\end{center}
\end{figure}

Similarly to motion planning methods, these joint motion and grasp planning methods can be categorized into sampling-based and optimization-based ones. Sampling-based methods~\cite{vahrenkamp2010integrated,fontanals2014integrated,haustein2017integrating} bias a motion planner to sample nodes that are closer to better grasps, and the grasps are synthesized online. The main limitation of these approaches is that the online synthesized grasps may not be accurate enough for precise grasping, especially for objects with complicated shapes. To overcome this limitation, several goal-set-based trajectory optimization methods are proposed~\cite{dragan2011manipulation,wang2019manipulation}. These methods first utilize an offline grasp planner to generate grasps of objects, where well-designed grasp planners can be used, such as grasps synthesized from physics simulation \cite{miller2004graspit,clemen2019}. These generated grasps are treated as goals in a goal set. The joint motion and grasp planner optimizes a collision-free trajectory that can reach one of the goals in the goal set. The goal set introduces a constraint on the last configuration of the robot trajectory. Using high-quality grasps as goals, these approaches can handle various objects in grasping.

In this work, motivated by the goal-set-based trajectory optimization framework for joint motion and grasp planning, we introduce a new trajectory optimization method for robotic grasping. Compared to previous methods~\cite{dragan2011manipulation,wang2019manipulation}, our method has the following advantages. First, we introduce a point-cloud representation of robots and task spaces for goal reaching and obstacle avoidance. Point clouds of robots are generated using the 3D meshes of the robot links, whereas point clouds of the task space can be obtained from depth sensors such as RGB-D cameras. Figure~\ref{fig:intro} shows the point cloud representation with the planned trajectories of a Fetch robot and a Franka Panda arm. This representation is general and can be used with any robot and any task space. Second, we formulate a constrained trajectory optimization problem using point-cloud representation for joint motion and grasp planning. Given a set of grasping goals, solving the optimization problem generates a trajectory to reach one of the goals that minimizes the objective function subject to certain constraints, such as joint limits. Instead of converting the constrained optimization problem into an unconstrained one and solving with first-order gradient descent-based techniques as in~\cite{dragan2011manipulation,wang2019manipulation}, we utilize the Interior Point OPTimizer (Ipopt)~\cite{wachter2006implementation} to solve the large-scale nonlinear optimization problem for trajectory planning, which can find better solutions compared to first-order solvers. Finally, we empirically verify our method on two robot grasping environments in the PyBullet simulator~\cite{pybullet}, i.e., a tabletop scene and a shelf scene, and demonstrate a significant improvement over the OMG-Planner~\cite{wang2019manipulation} in terms of metrics on grasping success and collision avoidance. In addition, we conducted real-world grasping experiments according to the SceneReplica benchmark~\cite{khargonkar2023scenereplica}. Our method improves over a sampling-based baseline in real-world experiments.

\vspace{-1mm}
\section{Related Work}
\vspace{-1mm}

\subsection{Manipulation Trajectory Optimization}

Trajectory optimization techniques have been successfully applied to robot manipulation. Early work such as CHOMP \cite{ratliff2009chomp} and related methods \cite{dragan2011manipulation,dragan2011learning} optimize a cost functional using covariant gradient descent. STOMP \cite{kalakrishnan2011stomp} uses stochastic sampling of noisy trajectories to optimize nondifferentiable costs. TrajOpt \cite{schulman2014motion} solves a sequential quadratic program, while GPMP2 \cite{mukadam2018continuous} formulates the problem as inference on a factor graph and finds the maximum a posteriori trajectory by solving a nonlinear least-squares problem. More recently, various trajectory optimization methods have been proposed to solve specific manipulation problems. For example, TORM~\cite{kang2020torm} is introduced to follow given end-effector paths. \cite{jin2021trajectory} solves a trajectory optimization problem for the manipulation of deformable objects. \cite{spahn2021coupled} solves a whole-body trajectory optimization for mobile manipulation. The advantage of trajectory optimization lies in its flexibility in introducing different cost functions and constraints for various problems. In this work, we solve a trajectory optimization problem for joint motion and grasp planning in robotic grasping.

\subsection{Joint Motion and Grasp Planning}

Traditionally, arm motion planning and grasp planning are tackled separately, which can result in suboptimal grasping trajectories. Since jointly optimizing trajectories and grasps is challenging, several approaches are proposed to solve a goal-constrained trajectory optimization problem for joint motion and grasp planning~\cite{dragan2011manipulation,wang2019manipulation,ichnowski2020gomp}, where grasps from a grasp planner such as GraspIt!~\cite{miller2004graspit} are used as goals. For example, \cite{dragan2011manipulation} projects the robot configuration of the last time step in the goal set during trajectory optimization.  OMG-Planner~\cite{wang2019manipulation} iterates between goal selection and trajectory optimization based on CHOMP. Recently, SE(3)-DiffusionFields~\cite{urain2023se} learns a cost function for grasp planning based on a diffusion model and then solved a joint optimization problem for grasp and motion planning. Unlike these methods, we introduce a cost function for goal reaching using our point-cloud representation and solve a constrained optimization problem for joint motion and grasp planning.

\section{Method}

\subsection{A Point-Cloud Representation for Robots and Task Spaces}

In robot motion planning, the goal is to generate a robot trajectory to reach a goal location while avoiding obstacles in the task space. The geometric representation of robots and the task space is a critical component of robot motion generation. A natural choice is to use 3D meshes of robots and objects in the task space. However, the limitation of using 3D meshes is that we cannot always obtain 3D meshes of objects, and collision checking between meshes is expensive. Another choice is to approximate robot links and obstacles in the task space with 3D shape primitives such as spheres, boxes, or cylinders~\cite{ratliff2009chomp,sundaralingam2023curobo}. Using 3D shape primitives simplifies collision checking, but results in inaccurate collision checking, where motion plans can be conservative. In this work, we utilize a simple geometric representation of objects and the task space, i.e., point clouds, for robot motion planning based on trajectory optimization.

\begin{figure} 
\centering
\begin{center} 
\includegraphics[width=0.48\textwidth]{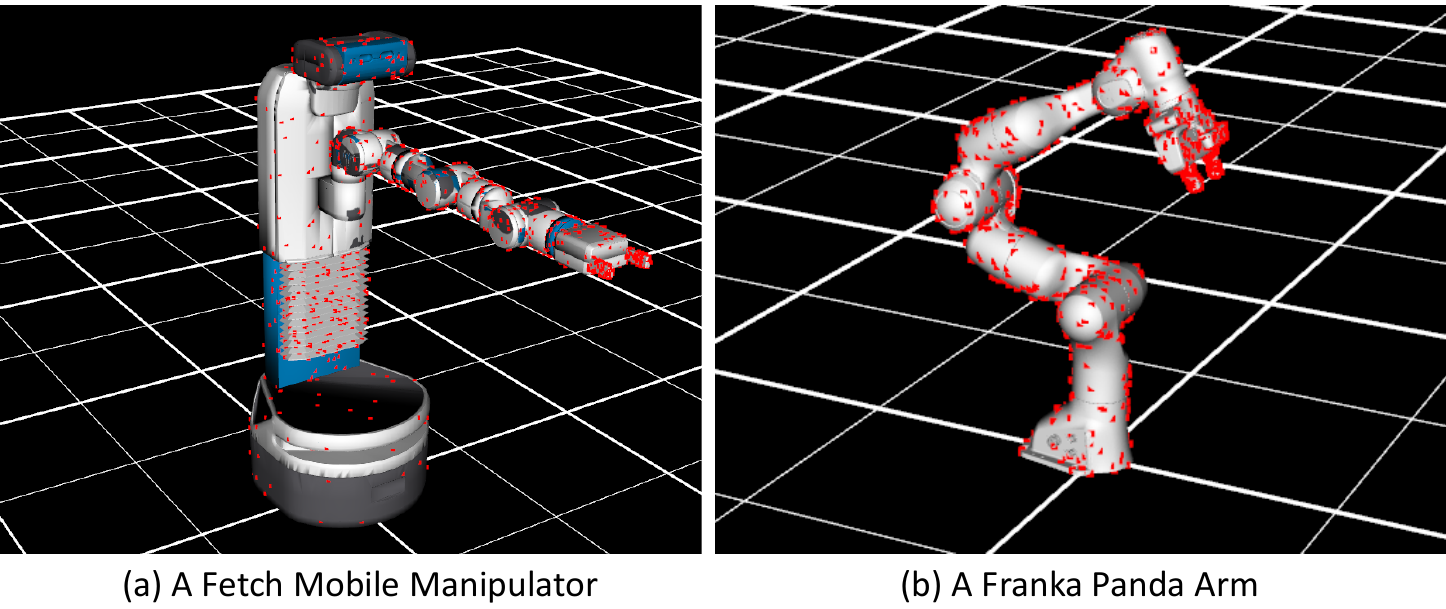}
\caption{Surface points (red points in the figure) are sampled as a representation for robots.}
\label{fig:robot_points}
\vspace{-6mm}
\end{center}
\end{figure}

\begin{figure*} 
\centering
\begin{center} 
\includegraphics[width=0.95\textwidth]{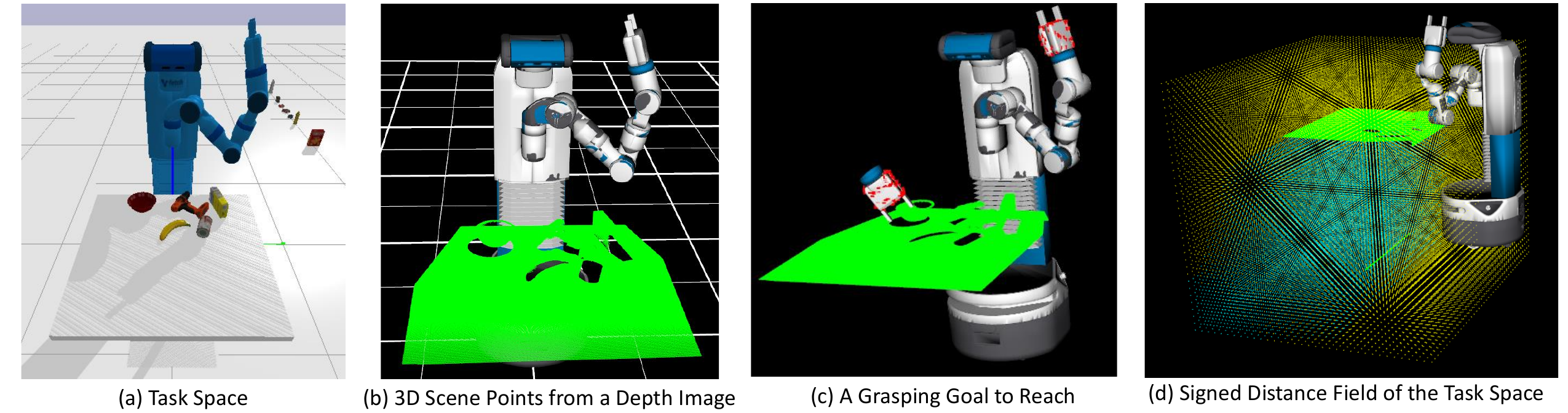}
\caption{(a) A tabletop scene for grasping with a Fetch robot. (b) A 3D point cloud of the scene computed using a depth image from the camera on the robot. (c) Reaching a grasping goal can be formulated as matching 3D points on the robot gripper. (d) Visualization of the signed distance field of the task space. Cyan points are with negative distances, and yellow points are with positive distances.}
\label{fig:task_points}
\vspace{-4mm}
\end{center}
\end{figure*}

Given a robot description using the Unified Robotics Description Format (URDF), each robot link has an associated 3D mesh model. To obtain a point-cloud representation of the robot, we simply sample 3D points from the vertices of the 3D meshes of the links. Figure~\ref{fig:robot_points} shows two examples of a Fetch mobile manipulator and a Franka Panda arm with their 3D points sampled, respectively. The number of points for each link is a parameter to set. Using more points requires more computation in goal reaching and obstacle avoidance, but it can achieve more accurate collision checking. We simply sample 100 points for each link in our experiments.

For objects in the robot task space, we cannot obtain 3D models of them if we want the robot to work in arbitrary environments. Therefore, we rely on depth sensing to obtain a point-cloud representation of the task space. By equipping a RGB-D camera with a robot, the robot can capture a depth image of the scene. Depth pixels can be back-projected to the camera frame using the intrinsic parameters of the camera. Then we can obtain a point cloud of the scene. Given the camera extrinsic parameters, i.e., 3D rotation and 3D translation of the camera in the robot base frame, the point cloud can be transformed into the robot base frame. Figure~\ref{fig:task_points}(a) shows a tabletop scene and a Fetch robot in the PyBullet simulator, and Figure~\ref{fig:task_points}(b) illustrates the computed point cloud using a depth image captured by the robot camera. Since RGB-D cameras are commonly used in robotic applications, using 3D scene points makes our approach generalizable to various scenarios. Next, we describe how to use the point-cloud representation in our grasping trajectory optimization method.


\subsection{Point Cloud-based Cost Function for Goal Reaching} \label{sec:cost_goal}

In grasping trajectory optimization, we need to generate a trajectory for a robot from its current joint configuration to a goal configuration for grasping a target object. The task-space goal is defined as an end-effector configuration to grasp the target object. For two-finger grippers, a goal can be simplified to be a homogeneous transformation $\mathbf{T}_g = (\mathbf{R}_g, \mathbf{t}_g) \in \mathbb{SE}(3)$, where $\mathbf{R}_g$ and $\mathbf{t}_g$ are the 3D rotation and the 3D translation of the gripper link with respect to the robot base frame, respectively.

In our method, we optimize for a trajectory that is discretized into $T$ time steps. The trajectory is parameterized by $T$ joint positions $\mathcal{Q} = (\mathbf{q}_1, \ldots, \mathbf{q}_T)$, where $\mathbf{q}_i, \in \mathbb{R}^n$ for $i=1, \ldots, T$, and $n$ is the degree of freedom of the robot. The last configuration of the trajectory $\mathbf{q}_T$ must reach the goal $\mathbf{T}_g$ in the task space. We can use forward kinematics to compute the end-effector pose of the robot at time step $T$: $\mathbf{T}(\mathbf{q}_T) = (\mathbf{R}_T, \mathbf{t}_T) \in \mathbb{SE}(3)$. We wish to define a cost function $c_\text{goal}(\mathbf{T}(\mathbf{q}_T), \mathbf{T}_g)$ to measure the distance between the gripper pose of the robot at the time step $T$ and the grasping goal. Consequently, minimizing this cost function can find a robot configuration to reach the goal.

Usually, the cost function is defined based on the distance between the two 3D rotations $(\mathbf{R}_T, \mathbf{R}_g)$ and the distance between the two 3D translations $(\mathbf{t}_T, \mathbf{t}_g)$. However, a weight must be adjusted to balance the two distances. Motivated by work on 6D object pose estimation~\cite{xiang2017posecnn}, we utilize the point matching loss function as our cost function for goal reaching. Let $\mathcal{E} = \{ \mathbf{x}_i \}_{i=1}^m$ be a set of $m$ 3D points on the end-effector of the robot (see Figure~\ref{fig:task_points}(c)). Our cost function for goal reaching is defined as
\begin{equation} \label{eq:cost_goal}
    c_\text{goal}(\mathbf{T}(\mathbf{q}_T), \mathbf{T}_g) = \sum_{i=1}^m\| (\mathbf{R}_T \mathbf{x}_i + \mathbf{t}_T) - (\mathbf{R}_g \mathbf{x}_i + \mathbf{t}_g)  \|^2,
\end{equation}
which minimizes the distance between two sets of point clouds undergone two homogeneous transformations. The advantage of using this cost function is that it eliminates the need to use a hyperparameter to balance rotation and translation. This cost function can also be generalized to grippers with high degrees of freedom, such as multi-finger grippers. Note that $\mathbf{T}(\mathbf{q}_T)$ is a function of $\mathbf{q}_T$ in the loss function according to forward kinematics.

\subsection{Point Cloud-based Cost Function for Collision Avoidance} \label{sec:cost_collision}

In addition to reaching the grasping goal, another requirement in robotic grasping is to avoid obstacles in the task space. We hope that the robot will not hit any object before grasping the target. For the example in Figure~\ref{fig:task_points}, the robot should avoid hitting the table and objects on the table during grasping. Instead of using 3D meshes or 3D shape primitives to represent obstacles, our method only has access to a point cloud of the scene. Therefore, we propose to compute a Signed Distance Field (SDF) of the robot task space using the point cloud for collision avoidance.

First, the extent of the task space is determined by the extent of the point cloud in the task space, where we add some margin to the point cloud space. Second, the SDF is constructed by densely sampling a 3D grid within the extent of the task space. The resolution of the grid is a parameter that can be tuned as a trade-off between computational efficiency and accuracy of collision checking. Third, we compute the signed distance value for each vertex of the 3D grid, which is approximated by the distance between the vertex and the closest point in the point cloud of the scene. The sign of the distance is determined by checking if the vertex is behind the point cloud or not. Specifically, we project the vertex to the depth image using the camera parameters and compare the depth values of the vertex and the projected pixel to obtain the distance sign. Figure~\ref{fig:task_points}(d) illustrates the SDF of the task space, where the cyan vertices have negative distances. Finally, using the computed SDF, we can check the collision between the robot and the scene by checking the signed distance values of the 3D points on the surface of the robot (Figure~\ref{fig:robot_points}) in the task space.

In addition, we can define a cost function for collision avoidance using the SDF. For each joint configuration in the robot trajectory $\mathbf{q}_t \in \mathbb{R}^n, t=1, \ldots, T$, let $\mathbf{x}(\mathbf{q}_t) \in \mathbb{R}^3$ be a surface point on the robot transformed into the task space according to the joint configuration $\mathbf{q}_t$ using forward kinematics. Then we can define a cost function for the 3D point $\mathbf{x}(\mathbf{q}_t)$ as in CHOMP~\cite{ratliff2009chomp}:
\begin{equation} \label{eq:cost_collision}
     c_\text{collision}(\mathbf{x}) =
    \begin{cases}
      -d(\mathbf{x}) + \frac{1}{2} \varepsilon & \text{if }  d(\mathbf{x}) < 0 \\
      \frac{1}{2 \varepsilon} ( d(\mathbf{x}) - \varepsilon )^2 & \text{if } 0 \leq d(\mathbf{x}) \leq \varepsilon \\
      0 & \text{otherwise}
    \end{cases},
\end{equation}
where $\varepsilon$ is a margin parameter and $d(\mathbf{x})$ is the signed distance of the 3D point. When the signed distance $d(\mathbf{x})$ is greater than $\varepsilon$, there is no cost in collision. Note that in our implementation, the SDF is precomputed using a 3D grid to speed up computation. Therefore, we simply find the voxel in which the 3D point $\mathbf{x}$ falls and use the signed distance value of the voxel as $d(\mathbf{x})$.

\subsection{Constrained Trajectory Optimization for Joint Motion and Grasp Planning}

With the designed cost functions for goal reaching and collision avoidance, we describe our trajectory optimization framework for joint motion and grasp planning. The task of a robot is to grasp a target object in a cluttered scene. We assume that there exists a grasp planner that can be used to synthesize grasps of the target. For example, in model-based grasping, GraspIt!~\cite{miller2004graspit} can be used to synthesize grasps given the 3D model of the target object. In model-free grasping, learning-based approaches such as 6DGraspNet~\cite{mousavian20196} or Contact-GraspNet~\cite{sundermeyer2021contact} can be used to synthesize grasps of the target object given the segmented point cloud of the target. We denote the set of synthesized grasps as a goal set $\mathcal{G} = \{ \mathbf{T}_i \}_{i=1}^K$, where $\mathbf{T}_i \in \mathbb{SE}(3)$ is a homogeneous transformation of the robot gripper and $K$ is the number of planned grasps. Our goal is to find a collision-free trajectory for the robot to reach one of the grasps.

We optimize for a trajectory that is discretized into $T$ time steps. The trajectory is parameterized by $T$ joint positions $\mathcal{Q} = (\mathbf{q}_1, \ldots, \mathbf{q}_T)$ and $T$ joint velocities $\dot{\mathcal{Q}} = (\dot{\mathbf{q}}_1, \ldots, \dot{\mathbf{q}}_T)$, where $\mathbf{q}_i, \dot{\mathbf{q}}_i \in \mathbb{R}^n$ for $i=1, \ldots, T$. Therefore, we optimize both the joint positions and the joint velocities in our method. Intuitively, we want to have the last joint position $\mathbf{q}_T$ reach one of the grasps in the goal set $\mathcal{G}$. Meanwhile, the trajectory should be collision-free and subject to constraints of the robot dynamics and joint limits. Formally, we solve the following constrained optimization problem to find the trajectory:
\begin{align} 
    \arg \min_{\mathcal{Q}, \dot{\mathcal{Q}} } & \Big( \min_{i=1}^K \big( c_{\text{goal}}(\mathbf{T}(\mathbf{q}_T), \mathbf{T}_i) + c_{\text{standoff}}(\mathbf{T}(\mathbf{q}_{T-\delta}), \mathbf{T}_i \mathbf{T}_{\Delta}) \big) \nonumber \\
    & + \lambda_1 \sum_{t=1}^T c_{\text{collision}}(\mathbf{q}_t) + \lambda_2 \sum_{t=1}^T \| \dot{\mathbf{q}_t} \|^2 \Big)  \label{eq:opt} \\
    \text{s.t.,} & \mathbf{q}_1 = \mathbf{q}_0  \\
    & \dot{\mathbf{q}}_1 = \mathbf{0}, \dot{\mathbf{q}}_T = \mathbf{0}  \\
    & \mathbf{q}_{t+1} = \mathbf{q}_t + \dot{\mathbf{q}}_t dt, t = 1, \ldots, T-1  \label{eq:dynamics}\\
    &  \mathbf{q}_l \leq  \mathbf{q}_t \leq  \mathbf{q}_u,  t = 1, \ldots, T  \label{eq:p_bound} \\
    & \dot{\mathbf{q}}_l \leq  \dot{\mathbf{q}}_t \leq  \dot{\mathbf{q}}_u,  t = 1, \ldots, T, \label{eq:v_bound}
\end{align}
where we minimize an objective function of $\mathcal{Q}$ and $\dot{\mathcal{Q}}$ subject to a set of constraints. Note that the objective function computes the minimum cost among all the grasping goals in the goal set $\mathcal{G}$. Consequently, solving the optimization problem will select the best goal from the goal set.

\begin{figure} 
\centering
\begin{center} 
\includegraphics[width=0.48\textwidth]{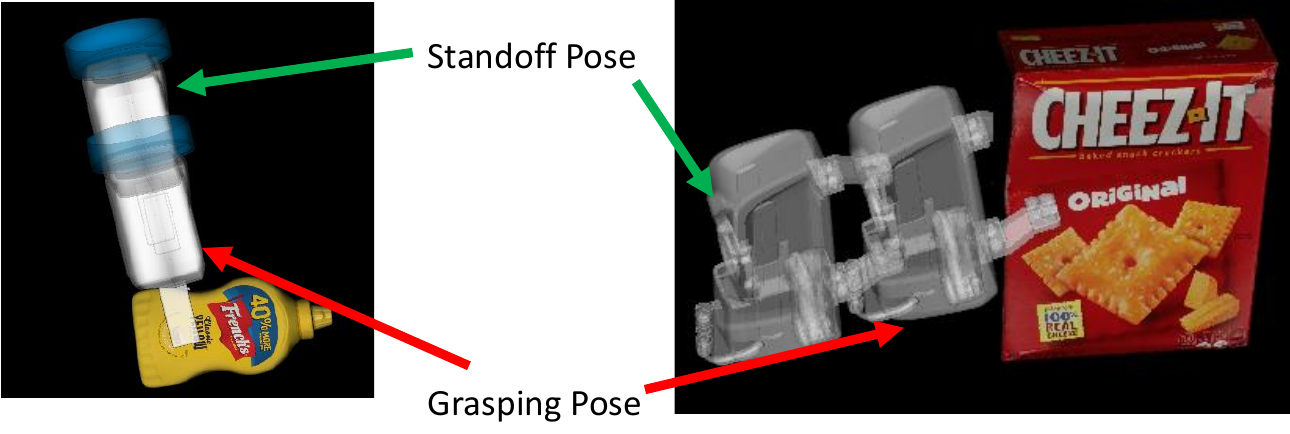}
\caption{Illustration of the grasping pose and the standoff pose for grasping.}
\label{fig:standoff}
\vspace{-6mm}
\end{center}
\end{figure}

\begin{figure*} 
\centering
\begin{center} 
\includegraphics[width=0.95\textwidth]{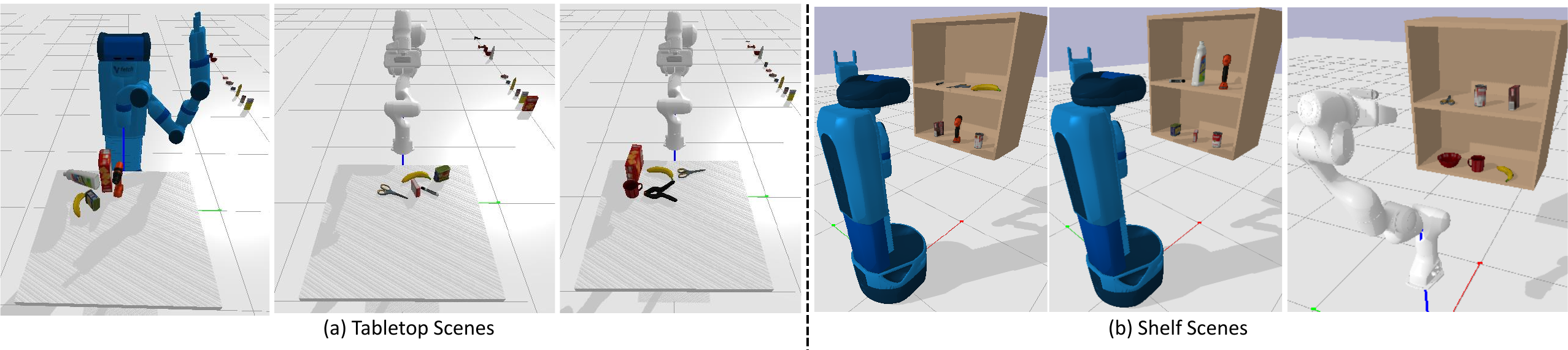}
\vspace{-2mm}
\caption{Examples of (a) tabletop scenes and (b) shelf scenes for grasping in PyBullet.}
\label{fig:scenes}
\vspace{-6mm}
\end{center}
\end{figure*}

First, the term $c_{\text{goal}}(\mathbf{T}(\mathbf{q}_T), \mathbf{T}_i)$ is the goal reaching cost described in Eq.~\eqref{eq:cost_goal} for the $i$th goal $\mathbf{T}_i$ in the goal set, where forward kinematics is used to compute the gripper pose given the robot configuration at the last time step $\mathbf{q}_T$. Second, in addition to reaching the goal in the last step, we introduce a cost term $c_{\text{standoff}}(\mathbf{T}(\mathbf{q}_{T-\delta}), \mathbf{T}_i \mathbf{T}_{\Delta})$ to ensure that the robot reaches a standoff pose for grasping before the goal. The standoff pose $\mathbf{T}_i \mathbf{T}_{\Delta}$ is computed by a displacement $\mathbf{T}_{\Delta} \in \mathbb{SE}(3)$ of the grasping pose $\mathbf{T}_i$ along the forward axis of the gripper as illustrated in Figure~\ref{fig:standoff}. The main reason of introducing the standoff pose is because optimizing the trajectory directly to reach the grasping pose may result in a collision between the robot and the target object. In these cases, the robot will knock down the target object and cannot grasp it. Adding the standoff pose in the trajectory optimization makes the problem simpler. In our objective function, we require that the robot gripper pose $\mathbf{T}(\mathbf{q}_{T-\delta})$ at time step $T-\delta$ to reach the standoff pose, where $\delta$ is a parameter to set. Finally, the objective function contains a cost term for collision avoidance and a cost term to penalize large velocities, where $\lambda_1$ and $\lambda_2$ are two weights to balance the costs. The collision cost for the time step $t$ is defined as
\begin{equation}
     c_{\text{collision}}(\mathbf{q}_t) = \sum_{i=1}^M  c_\text{collision}(\mathbf{x}_i(\mathbf{q}_t)),
\end{equation}
where $\mathbf{x}_i(\mathbf{q}_t) \in \mathbb{R}^3$ is a 3D point on the robot at the robot configuration $\mathbf{q}_t$ and $M$ is the total number of points on the robot. The collision cost is computed according to Eq.~\eqref{eq:cost_collision} using our SDF representation.

Next, we describe the constraints in the optimization problem. 1) $\mathbf{q}_1 = \mathbf{q}_0$, where $\mathbf{q}_0$ denotes the current configuration of the robot. This constraint ensures that the trajectory starts from the current configuration of the robot. 2) $\dot{\mathbf{q}}_1 = \mathbf{0}, \dot{\mathbf{q}}_T = \mathbf{0}$ ensure that the starting velocity and the ending velocity of the robot are zero. 3) $\mathbf{q}_{t+1} = \mathbf{q}_t + \dot{\mathbf{q}}_t dt$ ensures that the robot state follows the kinematics of the robot, where $dt$ is the time interval between two time steps. 4) The last two constraints in Eqs.~\eqref{eq:p_bound} and \eqref{eq:v_bound} ensure that the joint positions and the joint velocities are within the lower bounds $(\mathbf{q}_l, \dot{\mathbf{q}}_l)$ and upper bounds $(\mathbf{q}_u, \dot{\mathbf{q}}_u)$.

\subsection{Initialization for Grasping Trajectory Optimization}

The optimization problem in Eq.~\eqref{eq:opt} is a large-scale constrained nonlinear programming problem. For example, a Franka panda arm has $n=7$ DOFs. If we set the number of time steps of the trajectory $T=50$, the optimization problem has $7 \times 2 \times 50 = 700$ variables. We utilize the Interior Point OPTimizer (Ipopt)~\cite{wachter2006implementation} interfaced with the CasADi framework~\cite{Andersson2019} to solve it. Ipopt can only find local solutions that are sensitive to the initialization of the variables. To obtain a good local solution and speed up the optimization, we use the following strategy to initiate the optimization. 1) Given a set of grasping poses $\mathcal{G} = \{ \mathbf{T}_i \}_{i=1}^K$ of a target, we first filter out grasps that are in-collision with other objects in the scene. This collision checking can be achieved by checking the signed distance values of the 3D points on the robot gripper of a given pose as described in Section~\ref{sec:cost_collision}. 2) For the remaining grasps, we check if an inverse kinematics (IK) solution exists. We solve a simplified optimization problem to find an IK solution of a grasping goal $\mathbf{T}_{\text{g}}$:
\begin{align}  \label{eq:ik}
    \arg \min_{\mathbf{q}_T} \text{   } & c_{\text{goal}}(\mathbf{T}(\mathbf{q}_T), \mathbf{T}_{\text{g}})  \\
    \text{s.t.,}  &  \mathbf{q}_l \leq  \mathbf{q}_T \leq  \mathbf{q}_u,
\end{align}
where we use $\mathbf{q}_T$ to denote the variable in IK, and the objective function is the point matching cost function defined in~Eq.~\eqref{eq:cost_goal}. After finding a local solution $\mathbf{q}_T^*$, we compute the pose error between $\mathbf{T}(\mathbf{q}_T^*)$ and the goal $\mathbf{T}_g$ using a rotation error and a translation error. If both errors are smaller than some pre-defined thresholds, we claim that an IK solution is found. Otherwise, there is no IK solution for $\mathbf{T}_g$. In this way, we can filter out grasps without IK solutions. 3) For each remaining grasp with an IK solution, we interpolate a trajectory of the robot from the current configuration of the robot to the IK configuration. We then compute the collision cost of the trajectory $\sum_{t=1}^T c_{\text{collision}}(\mathbf{q}_t)$ to rank these trajectories. 4) Finally, we initialize the optimization with the trajectory that has the minimum collision cost. In the case of tie-breaking, e.g., multiple non-collision trajectories, we use the trajectory whose last configuration is closer to the current configuration of the robot. We empirically found that the above initialization process can speed up the convergence of the optimization to find a good local solution.

\section{Experiments}

We conducted experiments on 6DoF robotic grasping to evaluate our method in both simulation and in the real world. Two types of scenes are used for evaluation: a tabletop scene and a shelf scene as illustrated in Figure~\ref{fig:scenes} in the Pybullet simulator~\cite{pybullet}. In these scenes, 16 YCB objects~\cite{calli2015benchmarking} are used for grasping. The objects in the tabletop scenes are arranged according to the SceneReplica benchmark~\cite{khargonkar2023scenereplica}, and we sample object locations for the shelf scenes with 6 objects in each scene. Two robots, i.e., a Fetch mobile manipulator and a Franka Panda arm, are used for evaluation. The main evaluation metric is the success rate of grasping. If an object is successfully lifted by the robot, we count it as a success. In addition, we evaluate collisions during grasping.

\subsection{Implementation Details}

First, grasps of the 16 YCB objects are generated using GraspIt!~\cite{miller2004graspit}, with 100 grasps for each object. Therefore, the size of the goal set is 100. Second, the trajectory optimization is implemented based on the OpTaS library~\cite{mower2023optas}, which provides an interface to Ipopt solver using the CasADi framework~\cite{Andersson2019}. Third, the hyper-parameters in the method are set as follows. For each robot link, we sample 100 surface points. The margin $\varepsilon=0.02$ in computing the collision cost (Eq.~\eqref{eq:cost_collision}). The grid resolution of the signed distance field is 5cm. In the optimization problem Eq.~\eqref{eq:opt}, $\lambda_1=10$, $\lambda_2=0.01$ and $\delta=10$. The standoff pose for grasping is set as 10cm and 20cm from the grasping pose for the tabletop scenes and the shelf scenes, respectively. The number of time steps is $T=50$, and the time span of a trajectory is set to 10 seconds. Therefore, $dt=0.2$ in Eq.~\eqref{eq:dynamics}.

\begin{table}
\centering
\caption{Comparison between different loss functions for IK. Count is the total number of grasps tested for IK. The numbers of found IK solutions are presented for three loss functions.}
\vspace{-2mm}
\resizebox{0.48\textwidth}{!}
{
\begin{tabular}{l|c|ccc||c|ccc}
 & \multicolumn{4}{c||}{Tabletop (success $\uparrow$)} & \multicolumn{4}{c}{Shelf (success $\uparrow$)} \\
\hline

& \multicolumn{4}{c||}{Tabletop (success $\uparrow$ / collision $\downarrow$)} & \multicolumn{4}{c}{Shelf (success $\uparrow$ / collision $\downarrow$)} \\
\hline
 \multirow{2}{*}{Robot} & \multirow{2}{*}{Count} & Point  & Quater- & Euler & \multirow{2}{*}{Count} & Point  & Quater- & Euler \\
 & & matching & nion & angle & & matching & nion & angle \\ \hline


Fetch & 18,758  & \textbf{9,665}  & 9,600  & 8,288  &  10,917  & \textbf{2,911}  & 2,824 & 2,364   \\
Panda & 20,000 & \textbf{16,397}  &  15,905  & 10,596  & 12,000   & \textbf{3,925}  & 3,867  & 3,172  \\
                    \hline
\end{tabular}
}

\label{tab:ik}
\end{table}

\begin{table}
\centering
\caption{Statistics of grasping experiments in the PyBullet simulator}
\resizebox{0.48\textwidth}{!}
{
\begin{tabular}{l|c|ccc||c|ccc}
& \multicolumn{4}{c||}{Tabletop (success $\uparrow$ / collision $\downarrow$)} & \multicolumn{4}{c}{Shelf (success $\uparrow$ / collision $\downarrow$)} \\
\hline
\multirow{2}{*}{Object} & \multirow{2}{*}{Count} & Ours  & Ours & OMG~\cite{wang2019manipulation} & \multirow{2}{*}{Count} & Ours  & Ours & OMG~\cite{wang2019manipulation} \\
& & Fetch & Panda & Panda & & Fetch & Panda & Panda \\ \hline
cracker box  & 12 & 7 / 0  & 7 / 0  &  5 / 2  &  6  & 3 / 0  &  2 / 0 & 4 / 0
                   \\
sugar box & 10  &  10 / 0 &  10 / 0 & 10 / 0  &  5  & 4 / 0  & 5 / 0 & 4 / 0
                  \\
tomato soup can & 14 & 13 / 2  &  14 / 0 & 12 / 2 &  7  & 7 / 1 &  2 / 1 & 2 / 4
                     \\
mustard bottle &  14  &  12 / 0 & 11 / 0  & 11 / 0 &   5  &  3 / 0 &  4 / 0 & 4 / 0
                     \\
tuna fish can &  12  & 0 / 5  &  0 / 3  & 0 / 12  &   6  & 5 / 5  &  0 / 0  & 0 / 6
                     \\
pudding box &  10 &  7 / 0 &  6 / 0  &  6 / 2  &  7 & 5 / 1 &  4 / 0  & 2 / 3
                    \\
gelatin box & 14 & 11 / 0  & 10 / 0 &  4 / 2 &   7  &  6 / 1 &   1 / 1  & 2 / 3
                    \\
potted meat can &  14 &  10 / 0 & 14 / 0 &  12 / 0  &   10  & 8 / 0  &  10 / 0 & 6 / 1
                    \\
banana &  14  &  12 / 6 &  11 / 4 & 9 / 4  &  7  & 3 / 4  & 1 / 2  & 1 / 7
                    \\
bleach cleanser &  10 &  9 / 0 &  5 / 0  &  8 / 0 &   7  & 5 / 0  &  6 / 0  & 4 / 4
                    \\
bowl & 14         &  11 / 2 &  9 / 0 & 8 / 1 &   11 &  8 / 3  &  8 / 2  & 2 / 7
                    \\
mug & 10  &  8 / 1 &  4 / 1 & 6 / 0  &  10 & 3 / 4  &  6 / 2 &  5 / 4 \\   

power drill  & 14 &  14 / 2 & 12 / 0 &  12 / 0 &  7   & 4 / 0 & 2 / 1  & 1 / 4 \\

scissors &  14 &  2 / 0 &  1 / 6  & 1 / 13  & 11  &  5 / 7 &   1 / 5 & 0 / 10 \\

large marker & 12  &  1 / 3 &  4 / 5 & 4 / 9  &  5  & 3 / 2 & 0 / 3  & 0 / 5
                    \\
extra large clamp &  12  &  5 / 6 &  2 / 5   & 4 / 11  &  9  & 6 / 2  &  1 / 2  & 1 / 9
                    \\ \hline
ALL & 200 &  \textbf{132} / 27 &  120 / \textbf{24} &  112 / 58 &  120  & \textbf{78} / 30  &   53 / \textbf{19}  & 38 / 67 \\
                    \hline
\end{tabular}
}

\label{tab:grasping}
\vspace{-1mm}
\end{table}

\begin{figure*} 
\centering
\begin{center} 
\includegraphics[width=\textwidth]{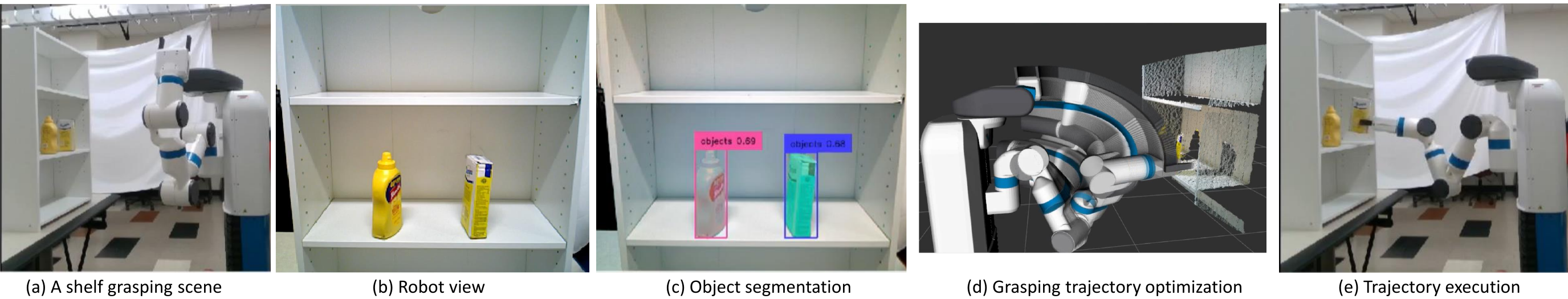}
\vspace{-4mm}
\caption{Illustration of the model-free grasping in the real world.}
\label{fig:model_free}
\vspace{-3mm}
\end{center}
\end{figure*}

\begin{table*}
\centering
\caption{Comparison between our grasping trajectory optimization (GTO) and the OMPL planning in~\cite{khargonkar2023scenereplica} for model-free grasping.}
\vspace{-3mm}
\resizebox{0.9\textwidth}{!}{
\begin{tabular}{c|c|c|c|c|ccc}
\hline
Method \# & Perception & Grasp Planning & Motion Planning & Control & Ordering & Pick-and-Place Success & Grasping Success  \\ \hline

\multicolumn{8}{c}{Model-free Grasping} \\ \hline

 1 &  MSMFormer~\cite{lu2022mean} &  Contact-graspnet~\cite{sundermeyer2021contact} + Top-down & OMPL~\cite{sucan2012open} & 
 MoveIt  & Near-to-far & 57 / 100  & 65 / 100\\ 

 2 &  MSMFormer~\cite{lu2022mean} &  Contact-graspnet~\cite{sundermeyer2021contact} + Top-down & \textbf{GTO (Ours)} & 
 MoveIt  & Near-to-far & \textbf{65 / 100}  & \textbf{71 / 100}\\

\hline
\end{tabular}
}
\label{tab:grasp_methods}
\vspace{-1mm}
\end{table*}

\begin{table}
\centering
\caption{Statistics of our grasping experiments for each YCB object. S: \#pick-and-place success, P$_{E}$F: \#perception failure, P$_L$F: \#planning failure, EF: \#execution failure}
\vspace{-1mm}
\resizebox{0.48\textwidth}{!}
{
\begin{tabular}{l|c|cccc|cccc} 
\hline
\multirow{2}{*}{Object} & \multirow{2}{*}{Count} & \multicolumn{4}{c|}{Method 1 (OMPL-based)} & \multicolumn{4}{c}{Method 2 (GTO-based Ours)} \\ \cline{3-10}
&  & S & P$_{E}$F & P$_{L}$F & EF & S & P$_{E}$F & P$_{L}$F & EF\\ \hline
\multicolumn{10}{c}{Order: Near-to-Far
} \\ \hline
cracker box  & 6 & 4  & 1 & - & 1 & 4  & - & - & 2
                   \\
sugar box & 5  &  5 & - & - & - &  5 & - & - & -
                  \\
 tomato soup can & 7 &  2 & 2 & 3 & - &  4 & 1 & 1 & 1
                     \\
mustard bottle & 7 &  6 & - & 1 & - &  2 & 1 & 3 & 1
                     \\
tuna fish can & 6  & 5  & 1 & - & - & 6  & - & - & -
                     \\
pudding box & 5 &  4 & 1 & - & - &  5 & - & - & -
                    \\
gelatin box & 7 & 6  & - & 1 & -  & 7  & - & - & -
                    \\
potted meat can & 7 &  5 & 2 & -  & - &  2 & 1 & 4  & -
                    \\
banana & 7  &  6 & - &  - &  1 &  7 & - &  - &  -
                    \\
bleach cleanser & 5 &  - & 1 & 2 & 2 & 2 & 1 & - & 2
                    \\
bowl & 7         &  6 & - & - & 1  &  7 & - & - & -
                    \\
mug & 5 & 2 & - & 2 & 1 & 2 & - & 3 & -
                    \\
scissors & 7 &  - & 2 &  2 & 3 &  3 & 3 &  - & 1
                    \\
power drill  & 7 &  3 & 3 & - & 1 &  2 & 3 & 1 & 1
                    \\
large marker & 6  &  1 & 2 & 2 & 1 &  3 & 1 & 2 & -
                    \\
extra large clamp & 6  &  2 & 1 & 2 & 1 &  4 & - & 2 & -
                    \\ \hline
ALL & 100 &  57 & 16 & 15 & 12 &  65 & 11 & 16 & 8 \\
                    \hline
\end{tabular}
}

\label{tab:grasp_object_metrics}
\vspace{-4mm}
\end{table}

\subsection{The Effect of Point Matching for Goal Reaching}

We evaluated the effectiveness of our point cloud-based representation for goal reaching. We solve the inverse kinematics optimization problem in Eq.~\eqref{eq:ik} with three different cost functions and compare their performance. The first one is the point-matching cost function in Eq.~\eqref{eq:cost_goal} to measure the difference between two transformations $\mathbf{T}_T$ and $\mathbf{T}_g$. For the other two cost functions, we use quaternions $(\Tilde{\mathbf{q}}_T, \Tilde{\mathbf{q}}_g)$ and Euler angles $(\mathbf{e}_T, \mathbf{e}_g)$ to represent the 3D rotations, and compute costs for rotation and translation separately:
\begin{align}
    c_\text{goal}^\text{quat}(\mathbf{T}_T, \mathbf{T}_g) &=  \| \mathbf{t}_T - \mathbf{t}_g \|^2 + 1 - (\Tilde{\mathbf{q}}_T \cdot \Tilde{\mathbf{q}}_g)^2, \\
    c_\text{goal}^\text{euler}(\mathbf{T}_T, \mathbf{T}_g) &=  \| \mathbf{t}_T - \mathbf{t}_g \|^2 + \| \mathbf{e}_T - \mathbf{e}_g \|^2,
\end{align}
where the distance between two quaternions measures the angular distance between the two rotations.

Using the three cost functions, we solve IK for each grasp of each object in the tabletop scenes and the shelf scenes. Table~\ref{tab:ik} presents the statistics of this experiment, where we count the number of successful IK solutions among all the trials. We consider an IK solution to be found after optimization if the translation error is less than 1 cm and the rotation error is less than 5 degrees. From the table, we can see that using the point matching cost function finds the maximum number of IK solutions, which validates the effectiveness our point-cloud based representation. Using distances between Euler angles is not a good choice due to the discontinuity between $-\pi$ and $\pi$.

\begin{figure*} 
\centering
\begin{center} 
\includegraphics[width=\textwidth]{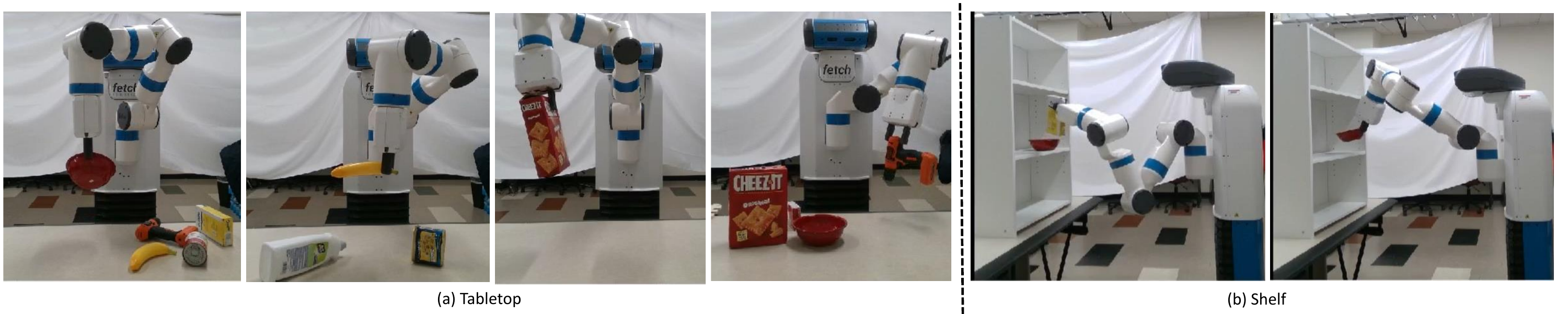}
\vspace{-6mm}
\caption{Examples of real-world grasping: (a) tabletop scenes and (b) shelf scenes}
\label{fig:results}
\vspace{-5mm}
\end{center}
\end{figure*}

\subsection{Simulation Results}

The results of our grasping experiments in PyBullet are presented in Table~\ref{tab:grasping}, where we compare our approach to the OMG-Planner~\cite{wang2019manipulation}. The OMG-Planner is a trajectory optimization method based on first-order gradient descent for joint motion and grasp planning. It alternates between goal selection and fixed-goal trajectory optimization. Grasps of the 16 YCB objects are generated using GraspIt!~\cite{miller2004graspit}. In the simulation, we query the object poses directly and then transform the grasps according to the object poses. We evaluate the number of successful grasps and the number of collisions during grasping. Using our point-cloud representation, we treat a grasping trajectory as in collision if there are 5 surface points of the robot with negative signed distances. 

For the table, we can see that 1) our method improves over the OMG-Planner in both the tabletop scenes and the shelf scenes. Our method achieves higher grasping success rates and lower collision rates. The main advantage of our method is that we solve a constrained nonlinear optimization problem with an advanced solver (Ipopt) compared to a gradient descent-based optimization. In addition, our point-cloud representation enables more accurate goal reaching and collision avoidance. 2) The Fetch robot achieves higher success rates compared to the Panda robot, largely due to its greater reachability and wider gripper. We cannot run the OMG-Planner for the Fetch robot since its implementation is tightly coupled with the Panda robot. In contrast, our implementation can be easily applied to different robots, where it only requires an URDF of a robot as input. 3) Some objects are more difficult to grasp. These are small or flat objects such as the tuna fish can, the scissors, the large marker, and the extra large clamp. Nonprehensile grasping strategies might be needed to grasp these objects successfully, which can be explored in future work.

\subsection{Model-free Grasping in the Real World}

Lastly, we conduct grasping experiments in the real world to evaluate our trajectory optimization method. We consider the task of model-free grasping, where we do not have 3D models of objects for perception and motion planning. Model-free grasping is applicable to diverse environments, and our approach does not rely on 3D object models. Figure~\ref{fig:model_free} illustrates the perception, planning, and control pipeline for model-free grasping.

We utilized the MSMFormer~\cite{lu2022mean} to segment unseen objects in an input RGB-D image for tabletop scenes. For shelf scenes, we found that MSMFormer cannot successfully segment objects in the shelf since it is not trained with similar scenes. Therefore, we used Grounding DINO~\cite{liu2023grounding} with text prompt ``objects'' to detect generic objects, and then used SAM~\cite{kirillov2023segment} to segment objects inside the bounding boxes from Grounding DINO. To synthesize grasps for a target object, we used Contact-GraspNet~\cite{sundermeyer2021contact}, which takes a segmented point cloud of an object as input and generates grasping poses of a parallel jaw gripper. These planned grasps are treated as goals in the goal set for joint motion and grasp planning. To execute a planned trajectory on a real robot, we also need to generate accelerations of the robot joints on the trajectory. Since our method does not solve for joint accelerations, we apply the path parameterization method~\cite{pham2018new} to reparameterize the planned trajectory.

We compare our method with an OMPL~\cite{sucan2012open}-based planning baseline in the SceneReplica benchmark~\cite{khargonkar2023scenereplica}. This baseline algorithm simply loops over all goals in the goal set and checks if there is a collision-free motion plan to reach a goal. The comparison results are presented in Table~\ref{tab:grasp_methods}. Our method achieves a better grasping success rate and a better pick-and-place success rate. Detailed evaluation statistics for each YCB object are presented in Table~\ref{tab:grasp_object_metrics}, where we classify pick-and-place failures into perception failures, planning failures, and execution failures. A detailed description of these failure types can be found in~\cite{khargonkar2023scenereplica}. Most failures are due to errors in object segmentation, grasp planning, and grasping goal selection. Because stable grasp is critical for pick-and-place success. By solving the trajectory optimization problem, our method benefits from better goal selection compared to the baseline algorithm. Figure~\ref{fig:results} shows some examples of successful grasping in the real world. Grasping videos can be found on the project page and in the supplementary material.

\subsection{Planning Time}

Our approach has demonstrated a significant improvement in planning efficiency over the OMPL-based baseline in the experiments conducted using the SceneReplica Benchmark \cite{khargonkar2023scenereplica}. On average, our method achieves a planning time of 15.4 seconds, which includes the computation time for the grasp collision checking, the IK checking, and the trajectory optimization. However, the OMPL-based baseline takes 45.6 seconds to find a grasp trajectory for a target object. In contrast, the OMG-Planner achieves 3.2 seconds planning time by solving parallel IKs and using GPUs for acceleration. We consider speeding up our method for future work.


\section{Conclusion and Discussion}

We introduce a new trajectory optimization method for joint motion and grasp planning. The core component of our method is a point cloud-based representation for robots and task spaces. This representation is generalizable to different robots and different environments. We formulate goal reaching and collision avoidance in the trajectory optimization using the point-cloud representation. By solving a constrained nonlinear optimization problem using the Ipopt solver, our method can generate robot trajectories for grasping. Experiments are conducted in simulation and in the real world to demonstrate the effectiveness of our method.

One limitation of our method is that trajectory optimization is slow when relying on an external solver. Future work includes speeding up the optimization. One direction is to explore using GPUs for parallel computing. Another direction is to explore model predictive control with our point-cloud representation for robotic grasping. To further improve the grasp success rate, a grasp planner that considers force closure or grasp stability will be helpful.

\noindent \textbf{Acknowledgement.} This work was supported in part by the DARPA Perceptually-enabled Task Guidance (PTG) Program under contract number HR00112220005 and the Sony Research Award Program. The work of T. Summers was supported by the United States Air Force Office of Scientific Research under Grant FA9550-23-1-0424 and the National Science Foundation under Grant ECCS-2047040.


\bibliographystyle{IEEEtran}
\bibliography{references}

\end{document}